\documentclass[10pt,conference]{IEEEtran}

\usepackage{cite}

\usepackage{url}
\usepackage{balance}
\usepackage[acronym]{glossaries}
\usepackage[table]{xcolor}
\usepackage{xspace}
\usepackage{booktabs}
\usepackage{multirow}
\usepackage[utf8]{inputenc}
\usepackage[caption=false]{subfig}
\usepackage[colorlinks=true,pagebackref=false,urlcolor=black,linkcolor=black,citecolor=black]{hyperref}

\usepackage[normalem]{ulem}

\newacronymstyle{long-short-br}
{%
  \GlsUseAcrEntryDispStyle{long-short}%
}%
{%
  \GlsUseAcrStyleDefs{long-short}%
}
\setacronymstyle{long-short-br}

\ifCLASSINFOpdf
   \usepackage[pdftex]{graphicx}
   \graphicspath{{figs/}}
   \DeclareGraphicsExtensions{.pdf,.jpeg,.png}
\else
   \usepackage[dvips]{graphicx}
   \graphicspath{{../figs/}}
   \DeclareGraphicsExtensions{.eps}
\fi

\begin{document}

\title{Simultaneous Iris and Periocular Region\\Detection Using Coarse Annotations
}

\newif\iffinal
\finaltrue
\newcommand{\cmtid}{75}


\iffinal


\author{Diego R. Lucio\IEEEauthorrefmark{1}, Rayson Laroca\IEEEauthorrefmark{1}, Luiz A. Zanlorensi\IEEEauthorrefmark{1}, 
Gladston Moreira\IEEEauthorrefmark{2}, David Menotti\IEEEauthorrefmark{1}\\
\IEEEauthorrefmark{1}Laboratory of Vision, Robotics and Imaging, Federal University of Paran\'a, Curitiba, PR, Brazil\\
\IEEEauthorrefmark{2}Laboratory of Intelligent Systems Computation, Federal University of Ouro Preto, Ouro Preto, MG, Brazil\\
\IEEEauthorrefmark{1}{\tt\small \{drlucio, rblsantos, lazjunior, menotti\}@inf.ufpr.br} \quad \IEEEauthorrefmark{2}{\tt\small gladston@iceb.ufop.br}
}


%

\else
  \author{Sibgrapi paper ID: \cmtid \\ }
\fi

\maketitle

\newacronym{asm}{ASM}{Active Shape Model}
\newacronym{asef}{ASEF}{Average of Synthetic Exact Filters}
\newacronym{leasm}{LE-ASM}{Local Eyebrow Active Shape Model}
\newacronym{cnn}{CNN}{Convolutional Neural Network}
\newacronym{encdec}{ED}{Encoder-Decoder}
\newacronym{fcn}{FCN}{Fully Convolutional Network}
\newacronym{gan}{GAN}{Generative Adversarial Network}
\newacronym{hog}{HOG}{Histogram of Oriented Gradients}
\newacronym{iou}{IoU}{Intersection over Union}
\newacronym{map}{mAP}{mean Average Precision}
\newacronym{iupui}{IUPUI}{IUPUI Multiwavelength}
\newacronym{masdv1}{MASD.v1}{Multi-Angle Sclera Dataset.v1}
\newacronym{nir}{NIR}{near-infrared}
\newacronym{prd}{PRD}{Periocular Region Detection}
\newacronym{relu}{Leaky ReLU}{Leaky Rectified Linear Unit}
\newacronym{vfc}{VFC}{Vector Field Convolution}

\newacronym{roi}{ROI}{Region of Interest}
\newacronym{svm}{SVM}{Support Vector Machines}
\newacronym{vj}{VJ}{Viola-Jones}
\newacronym{vis}{VIS}{visible}
\newacronym{rpn}{RPN}{Region Proposal Network}

\newcommand{\interval}{CASIA-Iris-Interval\xspace}
\newcommand{\distance}{CASIA-Iris-Distance\xspace}
\newcommand{\synth}{CASIA-Iris-Synt\xspace}
\newcommand{\warsaw}{LivDet-Iris 2015 Warsaw\xspace}
\newcommand{\clarkson}{LivDet-Iris 2015 Clarkson\xspace}
\newcommand{\berc}{BERC-Iris-Fake\xspace}

\newcommand{\lamp}{CASIA-Iris-Lamp\xspace}
\newcommand{\ar}{AR\xspace}
\newcommand{\mobbio}{MobBIO\xspace}
\newcommand{\thousand}{CASIA-Iris-Thousand\xspace}
\newcommand{\ubiris}{UBIRIS.v2\xspace}

\newcommand{\cross}{Cross-Eyed-VIS\xspace}
\newcommand{\csip}{CSIP\xspace}
\newcommand{\miche}{MICHE-I\xspace}
\newcommand{\niceii}{NICE-II\xspace}
\newcommand{\polyu}{PolyU-VIS\xspace}
\newcommand{\visob}{VISOB\xspace}

\newcommand{\faster}{Faster~R-CNN\xspace}
\newcommand{\yolo}{YOLOv2\xspace}
\newcommand{\mobbiofake}{MobBIOfake\xspace}

\newacronym{fast}{Fast R-CNN}{Fast Region-Based Convolutional Neural Network}
\newacronym{fpn}{FPN}{Feature Pyramid Network}
\newacronym{gs4}{MICHE-GS4}{MICHE Galaxy S4 Subset}
\newacronym{gt2}{MICHE-GT2}{MICHE Galaxy Tab 2 Subset}
\newacronym{ip5}{MICHE-IP5}{MICHE Iphone 5 Subset}
\newacronym{visob_iphone_day}{VISOB-IPHONE-DAY}{VISOB Iphone day light subset}
\newacronym{visob_iphone_dim}{VISOB-IPHONE-DIM}{VISOB Iphone dim light subset}
\newacronym{visob_iphone_office}{VISOB-IPHONE-OFFICE}{VISOB Iphone office light subset}

\newacronym{visob_oppo_day}{VISOB-IPHONE-DAY}{VISOB Iphone $5s$ day light subset}
\newacronym{visob_oppo_dim}{VISOB-IPHONE-DIM}{VISOB Iphone $5s$ dim light subset}
\newacronym{visob_oppo_office}{VISOB-IPHONE-OFFICE}{VISOB Iphone $5s$ office light subset}

\newacronym{visob_oppo_day}{VISOB-OPPO-DAY}{VISOB Oppo $N1$ day light subset}
\newacronym{visob_oppo_dim}{VISOB-OPPO-DIM}{VISOB Oppo $N1$ dim light subset}
\newacronym{visob_oppo_office}{VISOB-OPPO-OFFICE}{VISOB Oppo $N1$ office light subset}

\newacronym{visob_samsung_day}{VISOB-SAMSUNG-DAY}{VISOB Galaxy note $4$ day light subset}
\newacronym{visob_samsung_dim}{VISOB-SAMSUNG-DIM}{VISOB Galaxy note $4$ dim light subset}
\newacronym{visob_samsung_office}{VISOB-SAMSUNG-OFFICE}{VISOB Iphone Galaxy note $4$ light subset}

\newacronym{iiitdcli}{IIIT-D CLI}{IIIT-Delhi Contact Lens Iris}

\newacronym{ndcld15}{NDCLD15}{Notre Dame Contact Lens Detection 2015}

\newacronym{mobbio}{MobBIO}{MobBIO Subset}

\newacronym{ndccl}{NDCCL}{Notre Dame Cosmetic Contact Lenses}

\newacronym{frgc}{FRGC}{Face Recognition Grand Challenge}
\newacronym{mbgc}{MBGC}{Multiple Biometrics Grand Challenge}
\newacronym{mrf}{MRF}{Markov Random Field}

\newcommand{\supplementary}{https://web.inf.ufpr.br/vri/databases/iris-periocular-coarse-annotations/}
\begin{abstract}
In this work, we propose to detect the iris and periocular regions simultaneously using coarse annotations and two well-known object detectors: \yolo and \faster.
We believe coarse annotations can be used in recognition systems based on the iris and periocular regions, given the much smaller engineering effort required to manually annotate the training images.
We manually made coarse annotations of the iris and periocular regions ($\approx$~122K images from the \gls*{vis} spectrum and $\approx$~38K images from the \gls*{nir} spectrum).
The iris annotations in the \gls*{nir} databases were generated semi-automatically by first applying an iris segmentation~CNN and then performing a manual inspection.
These annotations were made for 11 well-known public databases (3~\gls*{nir} and 8~\gls*{vis}) designed for the iris-based recognition problem, and are publicly available to the research community\footnote{All annotations made by us are publicly available at the following website: \supplementary.}.
Experimenting our proposal on these databases, we highlight two results. First, the Faster R-CNN + \gls*{fpn} model reported an \gls*{iou} higher than \yolo (91.86\%~vs~85.30\%).
Second, the detection of the iris and periocular regions being performed simultaneously is as accurate as performed separately, but with a lower computational cost, i.e., two tasks were carried out at the cost of one.
\end{abstract}

\IEEEpeerreviewmaketitle

\section{Introduction}

\glsresetall

In recent years, the interest in biometrics to automatically identify and/or verify a person's identity has greatly increased~\cite{abdelwhab2018survey, bowyer2016handbook}. Biometrics refers to the use of physiological and behavioral characteristics of humans for personal identification~\cite{das2014sclera}. 
Such characteristics are particularly important since they cannot be changed, forgotten, lost or stolen, providing an unquestionable connection between the individual and the application that makes use of them~\cite{menotti2015deep}.

Several characteristics of the human body can be used as biometrics such as fingerprints, face, ocular region components and voice, each with advantages and disadvantages.
Among the aforementioned modalities, ocular biometric traits have received significant attention in the recent past~\cite{das2014, NIGAM20151, lucio2018fully} due to the fact that the ocular region is an important and interrelated human trait consisting of several parts, for example, the cornea, lens, optic nerve, retina, pupil, iris, and periocular region.  
In this direction, many authors proposed biometric systems based on iris, periocular, retina, and sclera regions, as they are considered potential biometric modalities~\cite{retina_patent,jain}. 

The iris appears as one of the main biological characteristics in security systems since it remains unchanged over time and its uniqueness level is high~\cite{zhu2000biometric}. 
Furthermore, the identification using the iris region is non-invasive, that is, there is no need for physical contact to obtain and analyze an iris image~\cite{jain2006biometrics}.
However, after decades of research in personal identification, it has been observed that better results can be achieved by combining different biometric modalities~\cite{NIGAM20151,tan_and_kumar,chang2004multi}.
A good example of it is the combination of iris and periocular-based biometrics~\cite{xiao2012fusion,DeMarsico2017}.

In this work, we compare the detection of the iris and periocular regions being performed separately and simultaneously using two well-known object detection networks: \yolo~\cite{redmon:2017} and \faster~\cite{faster}.
Such deep models were chosen due to the fact that (i)~promising results were recently obtained using them in other detection tasks~\cite{faster_ref2, faster_ref1, laroca2018robust}; and (ii)~handcrafted features are easily affected by noise and might not be robust for unconstrained scenarios.

Typically, in biometric systems that use iris and/or periocular region images as input, the first step in which efforts should be applied is the detection of the \gls*{roi}~\cite{daugman1993high}, as a poor detection would probably impair the effectiveness of the subsequent steps of the system~\cite{daugman2004iris,tan_and_kumar}.
Recently, Zanlorensi et~al.~\cite{zanlorensi2018impact} showed that impressive iris recognition rates can be achieved when using deep representations having as system input the bounding boxes of the iris region, without the iris segmentation preprocessing.
Also using deep representations and having as input a squared region (i.e., a bounding box), Luz et~al.~\cite{luz:2018} achieved state-of-the-art results for periocular recognition.
Such results, shorter execution times compared to single detection approaches (in which the iris and the periocular region are detected separately), and the promising results obtained in preliminary experiments support our motivation to detect both regions simultaneously.

The main contributions of this paper can be summarized as follows:
(i)~two new approaches for the simultaneous detection of the iris and periocular region;
(ii)~a comparative evaluation between detecting both regions simultaneously or separately in \textbf{eleven} publicly available databases; and
(iii)~for learning the models used in the experiments, coarse annotations (i.e., bounding boxes) were manually made for both iris and periocular regions of $122{,}738$ images from $8$ well-known \gls*{vis} spectral databases.
As stated by Cordts et al.~\cite{cordts2016cityscapes}, coarse annotations are intended to support research areas that exploit large volumes of data.
We also automatically generated $38{,}851$ bounding boxes using the iris segmentation approach proposed by Bezerra~et~al.~\cite{bezerra2018robust} for $3$ well-known \gls*{nir} spectral databases. We manually checked and corrected (if necessary) all annotations.

We chose the approach proposed in~\cite{bezerra2018robust} due to the fact that it presented an error rate lower than $1.5$\% in the aforementioned \gls*{nir} databases. However, despite the good results presented by that segmentation approach, the detection task is much less expensive in terms of both computational cost and data annotation.
Regarding the $11$ databases employed in our experiments, they were chosen because they are widely used in the biometric recognition literature~\cite{zanlorensi2018impact, silva2018multimodal, aginako2016machine,deshpande2014iris}, which we plan to investigate in future works.
It should be noted that, in many works in the literature, no more than three databases were used in the experiments~\cite{rodriguez2005new,zhang2014new, alvarez2010fast, zhou2013new}.

In our experiments, the \faster model yielded \gls*{iou} values higher than \yolo ($91.86$\% vs $85.30$\%) and the detection of the iris and periocular regions being performed simultaneously is as accurate as performed separately, but with a lower computational cost, i.e., two tasks were carried out at the cost of one.
Regarding the use of coarse annotations, we believe they can be used in recognition systems based on the iris and periocular regions, given the much smaller engineering effort required to manually annotate the training~images.

The remainder of this paper is organized as follows. 
We review related works in Section~\ref{sec:related_work}.
In Section~\ref{sec:proposed_approach}, our methodology is described. 
Section~\ref{sec:experiments} and Section~\ref{sec:results} present, respectively, the experimental setup and the results obtained.
Finally, conclusions and future work are discussed in Section~\ref{sec:conclusions}.
\section{Related Work}
\label{sec:related_work}

In this section, we discuss works related to iris and periocular region detection and conclude with final~remarks.

\subsection{Iris Detection}
\begin{figure}[!htb]
	\vspace{-3.5mm}
	\centering
	\subfloat[Rectangular bounding box\label{fig:squared_bounding_box}]{\includegraphics[width=0.45\columnwidth]{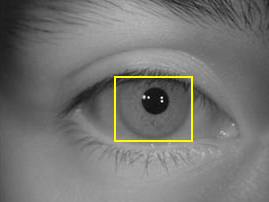}
	}%
	\subfloat[Outer iris contour\label{fig:outer_iris}]{\includegraphics[width=0.45\columnwidth]{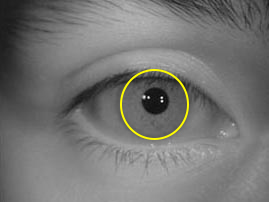}
	}%
	\caption{Samples of representation to iris~\gls*{roi} extraction.}%
	\vspace{-1mm}
\end{figure}

Regarding iris detection, the works in the literature commonly show the detected \gls*{roi} using two different representations. Fig.~\ref{fig:squared_bounding_box} shows the use of a rectangular bounding box as the iris delimitation, while Fig.~\ref{fig:outer_iris} shows an elliptical \gls*{roi} detection using the outer iris boundary. 

Many works in the literature show the iris delimitation by using an elliptical contour around the outer edge of it. Daugman~\cite{daugman1993high} pioneered this scenario by proposing an approach that makes the use of an integro-differential operator to detect the iris identifying the borders present in the images.
This operator takes into account the circular shape of the iris to find its correct position by maximizing the partial derivative concerning the radius.
In the experiments, the author employed a private database composed of $592$ eye images captured in the \gls*{nir} wavelength from $323$ subjects.

Zhang \& Ma~\cite{zhang2014new} adopted a method that employs a momentum-based level set~\cite{lathen2009momentum, wang2010momentum} along with the Daugman's operator to locate the pupil boundary.
Specifically, an initial contour of the iris is obtained with a momentum-based level set using the minimum average gray level.
Then, the integro-differential operator is applied to perform the final detection, reducing the execution time and improving the results obtained in~\cite{daugman1993high}. An accuracy rate of $98.53\%$ was achieved on the CASIA-IrisV2 database~\cite{CASIA2004}. Such improvement occurs because the initially detected contour is generally close to the actual inner boundary of the~iris.

Alvarez-Betancourt \& Garcia-Silvente~\cite{alvarez2010fast}, on the other hand, presented an iris location method based on the detection of circular boundaries through gradient analysis in points of interest of successive arcs, reaching an accuracy of $98\%$ on the CASIA-IrisV3 database~\cite{CASIA2010} with improvements in processing time.
The quantified majority operator QMA-OWA, proposed in~\cite{pelaez2006majority}, was used to obtain a representative value for each successive arc. Then, the iris boundary is given by the arc with the most significant representative~amount.

In the method proposed by Cui~et~al.~\cite{cui2012rapid}, the eyelashes are removed as a first step using the dual-threshold method, which can be an advantage over other iris location approaches.
Next, the facula is removed by using mathematical morphology. Finally, the accurate iris position is obtained through Hough-Transform and least-squares. Their method achieved $ 98$\% accuracy in the CASIA-IrisV3-Twins database~\cite{CASIA2010}.

Zhou~et~al.~\cite{zhou2013new} presented a method for iris location in which the initial position of the iris is obtained by using the~\gls*{vfc} technique.
This initial estimate makes pupil location much closer to the actual boundary instead of circle fitting, improving location accuracy and reducing computational cost.
The final result is obtained using the algorithm proposed by Daugman, reducing the computational cost and improving the location accuracy since the pupil delineation is much closer to the actual boundary.
An accuracy rate of $98.85$\% was reported on the CASIA-IrisV2~database.

Su~et~al.~\cite{su2017iris} proposed an iris location algorithm based on local property and iterative searching, achieving $98.08\%$ accuracy on the CASIA-IrisV1 and CASIA-IrisV3 databases (i.e., they were combined in their experiments).
In order to detect the \gls*{roi}, the pupil area is extracted using iris regional attributes, and the inner edge of it is fitted by iterating, comparing and sorting the pupil edge points. The outer edge location is made by using an iterative searching method from the extracted pupil center and radius, with a shorter time in relation to the approaches available in the literature.

Chen~\&~Ross~\cite{chen2018multitask} designed a multi-task \gls*{cnn}-based approach for joint iris and presentation attack detection. The experiments were performed on six publicly available databases, however, iris detection results were not reported as the main focus of their work is to identify presentation~attacks.

Severo~et~al.~\cite{severo2018benchmark} represented the iris as a rectangular bounding~box. 
They fine-tuned the Fast-YOLOv2 model, which is much faster but less accurate than \yolo, in order to perform the \gls*{roi} extraction, overcoming problems such as noise, eyelids, eyelashes and reflections.
Six public databases were used to evaluate their method, which attained accuracy rates above $97$\% in all of~them.

Wang~et\cite{wang2019joint} recently introduced IrisParseNet, a network for iris detection that reached $89.40$\%, $85.39$\% and $85.07$\% \gls*{iou} values in the \distance, \ubiris and \miche databases, respectively.
Their method simultaneously estimates the pupil center, the iris segmentation mask, and the iris inner/outer boundaries.

\subsection{Periocular Region Detection}

Park~et~al.~\cite{Park2011} proposed one of the first biometric approaches based on the periocular region, featuring an eye region detector that uses face images detected by the Viola-Jones detector~\cite{Viola2004} as input and outputs the periocular~region.

Similarly, Juefei-Xu~\&~Savvides~\cite{juefei2012} also proposed a periocular region detection approach that employs as input a face image detected by the Viola-Jones detector.
Nevertheless, the periocular region is identified using \glspl*{asm} that identify $79$ facial landmarks, containing points relative to the eye region among them.

Mahalingam~et~al.~\cite{mahalingam_2014} designed an eye detector that receives a face image and outputs the periocular region through \gls*{asef}. All experiments were carried out on a private database composed of $1.2$~million faces from $38$~subjects.
Le~et~al.~\cite{le_2014}, on the other hand, proposed a \gls*{leasm} to first detect the eyebrow region directly from a given face image and then to detect the periocular region using \glspl*{asm}.
The results obtained on this particular stage (i.e., periocular region detection) were not~reported.

Proença~et~al.~\cite{proenca_2014} proposed a \gls*{mrf} method to segment the periocular region components and other elements around them~(i.e., the iris, sclera, eyelashes, eyebrows, hair, skin and glasses).
Their approach analyzes the image pixels and outputs the segmented region taking into account appearance and geometrical constraints and assuring that the system output is biologically plausible.
The periocular region can be predicted by combining the outer limits of the sclera and the lower eyelashes.

\subsection{Final Remarks}

In most works, the accuracy was employed as the evaluation metric for iris and periocular region detection. However, the authors used different protocols to calculate the accuracy or do not specifically describe how the accuracy obtained by their approach was computed. Therefore, it is plausible to question how robust one method is compared to another.

While in the iris detection scenario a poor description of the evaluation metrics used has been made, in the periocular region detection scenario none of the studies found in the literature report the results achieved in this particular stage, probably due to the fact that the detection of the \gls*{roi} was considered only as a preprocessing step in such works~\cite{Park2011,juefei2012, mahalingam_2014, le_2014}.

Taking this information into consideration and also the fact that \glspl*{cnn} are not widely explored in the iris and periocular region detection domains, we propose to evaluate two well-known~\gls*{cnn} object detectors (i.e., \yolo and \faster) in \textbf{eleven} coarsely annotated databases.

More specifically, the main objective of this work is to evaluate the simultaneous detection of the iris and periocular regions.
The simultaneous detection approach is proposed taking into account the assumption that \glspl*{cnn} are able to understand the context present in the images, thus improving the results obtained by conventional single detection approaches.
\section{Methodology}
\label{sec:proposed_approach}

Currently, one of the most accurate ways to perform image classification, segmentation and object detection is using deep \glspl*{cnn}.
Therefore, in this work, we propose the simultaneous detection of the iris and periocular regions using two object detection models: \yolo~\cite{redmon:2017} and \faster~\cite{faster}.
It should be noted that (i)~we trained both models from scratch; 
(ii)~such models were chosen because promising results were obtained using them in other detection tasks~\cite{faster_ref2, faster_ref1, laroca2018robust}.

Our hypothesis is that the proposed simultaneous detection approach is able to understand the context of the image and thereby improve detection results compared to single detection approaches in which the iris and the periocular region are detected separately.
As baselines, we also adopted the \yolo and \faster models, but in two independent detection steps, i.e., one for the iris and one for the periocular~region.

\subsection{\yolo}

Table~\ref{tab:yolo} presents the \yolo model, employed for detecting the iris and the periocular region.
The architecture has $19$ convolutional and $5$ max-pooling layers. The convolutional layers, except for the last one, are divided into two groups: external and internal. The layers belonging to the external group use kernels of size $3\times3$, whereas the layers belonging to the internal group use kernels of size $1\times1$. Alternating $1\times1$ convolutional layers reduce the features space from preceding layers~\cite{redmon2016yolo}. The convolutional blocks are composed of: convolution, batch normalization, and a \gls*{relu}.

As this model does not have fully connected layers, it can receive images of any size as input. We adopted an input size of $416\times416$ pixels due to the good results achieved employing these dimensions in~\cite{redmon:2017}. 
We also reduced the number of filters in the last convolutional layer to match our number of classes.
The number of filters in that layer is given by
\begin{equation}
	\label{eq:filters_number}
	\textit{filters} = (C + 5) \times A \, ,
\end{equation}
\noindent where $A$ is the number of anchor boxes (we use $A$~=~$5$) used to predict bounding boxes and $C$ is the number of classes, in our case either $C=1$ or $C=2$ to detect the iris and periocular regions separately or simultaneously, respectively.
Thus, there are $30$ filters in the last convolutional layer when the regions are detected separately and $35$ when they are detected simultaneously.

The main difference between the \yolo model proposed in~\cite{redmon:2017} and the one used in this work is that we removed the route layers, i.e., layers that concatenate a list of previous layers together. In preliminary experiments, we observed that removing such layers did not negatively affect the results obtained in our tasks and also reduced the execution~time.

\begin{table}[!htb]
	\centering	
    \caption{The YOLOv2 model, modified for the detection of the iris and the periocular region. There are $30$ filters in the last convolutional layer when the regions are detected separately and $35$ when they are detected simultaneously.}
	\label{tab:yolo}	
	\vspace{-1.5mm}
	\resizebox{.98\columnwidth}{!}{	
	\begin{tabular}{@{ }ccccccc@{ }}
		\toprule
		\textbf{\#} & \textbf{Layer} & \textbf{Group} & \textbf{Filters} & \textbf{Size} & \textbf{Input} & \textbf{Output} \\ \midrule
		$0$ & conv & External & $32$ & $3 \times 3 / 1$ & $416 \times 416 \times 1 / 3$ & $416 \times 416 \times 32$ \\
		$1$ & max &  &  & $2 \times 2 / 2$ & $416 \times 416 \times 32$ & $208 \times 208 \times 32$ \\
		$2$ & conv & External & $64$ & $3 \times 3 / 1$ & $208 \times 208 \times 32$ & $208 \times 208 \times 64$ \\
		$3$ & max & &  & $2 \times 2 / 2$ & $208 \times 208 \times 64$ & $104 \times 104 \times 64$ \\
		$4$ & conv & External & $128$ & $3 \times 3 / 1$ & $104 \times 104 \times 64$ & $104 \times 104 \times 128$ \\
		$5$ & conv & Internal & $64$ & $1 \times 1 / 1$ & $104 \times 104 \times 128$ & $104 \times 104 \times 64$ \\
		$6$ & conv & External & $128$ & $3 \times 3 / 1$ & $104 \times 104 \times 64$ & $104 \times 104 \times 128$ \\
		$7$ & max & &  & $2 \times 2 / 2$ & $104 \times 104 \times 128$ & $52 \times 52 \times 128$ \\
		$8$ & conv & External & $256$ & $3 \times 3 / 1$ & $52\times 52 \times 128$ & $52 \times 52 \times 256$ \\
		$9$ & conv & Internal & $128$ & $1 \times 1 / 1$ & $52 \times 52 \times 256$ & $52 \times 52 \times 128$ \\
		$10$ & conv & External & $256$ & $3 \times 3 / 1$ & $52 \times 52 \times 128$ & $52 \times 52 \times 256$ \\
		$11$ & max & &  & $2 \times 2 / 2$ & $52 \times 52 \times 256$ & $26 \times 26 \times 256$ \\
		$12$ & conv & External & $512$ & $3 \times 3 / 1$ & $26 \times 26 \times 256$ & $26 \times 26 \times 512$ \\
		$13$ & conv & Internal& $256$ & $1 \times 1 / 1$ & $26 \times 26 \times 512$ & $26 \times 26 \times 256$ \\
		$14$ & conv & External & $512$ & $3 \times 3 / 1$ & $26 \times 26 \times 256$ & $26 \times 26 \times 512$ \\
		$15$ & conv & Internal& $256$ & $1 \times 1 / 1$ & $26 \times 26 \times 512$ & $26 \times 26 \times 256$ \\
		$16$ & conv & External& $512$ & $3 \times 3 / 1$ & $26 \times 26 \times 512$ & $26 \times 26 \times 512$ \\
		$17$ & max & &  & $2 \times 2 / 2$ & $26 \times 26 \times 512$ & $13 \times 13 \times 512$ \\
		$18$ & conv & External & $1024$ & $3 \times 3 / 1$ & $13 \times 13 \times 512$ & $13 \times 13 \times 1024$ \\
		$19$ & conv & Internal & $512$ & $1 \times 1 / 1$ & $13 \times 13 \times 1024$ & $13 \times 13 \times 512$ \\
		$20$ & conv & External & $1024$ & $3 \times 3 / 1$ & $13 \times 13 \times 512$ & $13 \times 13 \times 1024$ \\
		$21$ & conv & Internal & $512$ & $1 \times 1 / 1$ & $13 \times 13 \times 1024$ & $13 \times 13 \times 512$ \\
		$22$ & conv & External & $1024$ & $3 \times 3 / 1$ & $13 \times 13 \times 512$ & $13 \times 13 \times 1024$ \\
		$23$ & conv & & $30/35$ & $1 \times 1 / 1$ & $13 \times 13 \times 1024$ & $13 \times 13 \times 30/35$ \\
		$24$ & detection & &  &  &  &  \\ \bottomrule
	\end{tabular} \,
	}
	
\end{table}

\subsection{Faster R-CNN + Feature Pyramid Network}

\begin{figure*}[!htb]
	\centering
	\includegraphics[width=0.95\linewidth]{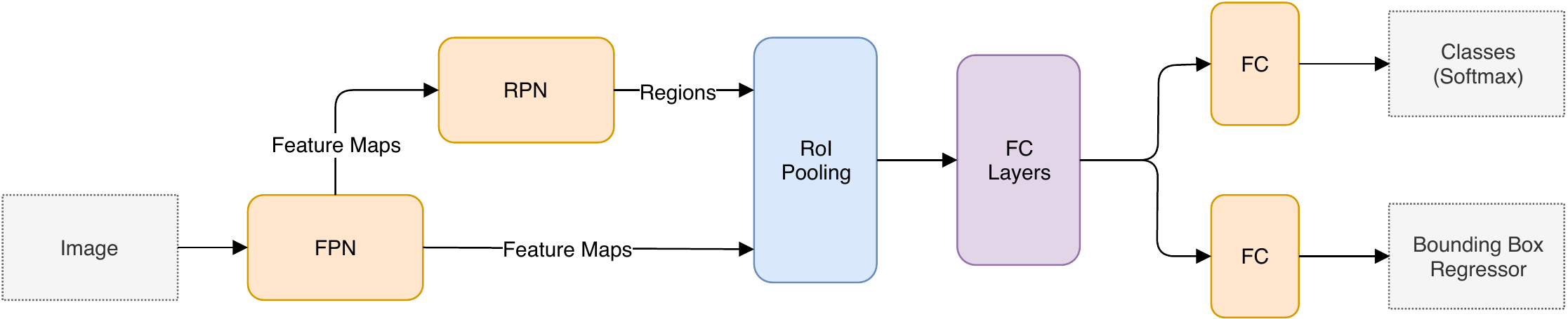}
	\caption{\faster + \gls*{fpn} architecture overview.}
	\label{fig:pyramid}
	\vspace{-1mm}
\end{figure*}

We employ the \faster model~\cite{faster} combined with a \gls*{fpn}~\cite{FPN}, as shown in Figure~\ref{fig:pyramid}.
\faster is commonly composed of (i)~a feature map extraction network; (ii)~a region proposal network and (iii)~a detection network.
We replaced the standard \gls*{cnn} feature extraction module by an \gls*{fpn}, and thus multiple feature map layers are generated with better quality information than the regular implementation of~\faster. 

\subsection{Coarse Annotations}

\begin{figure}[!htb]
	\vspace{-3mm}
	\centering
	\subfloat[Fine\label{fig:masks_original:a}]{\includegraphics[width=0.45\columnwidth]{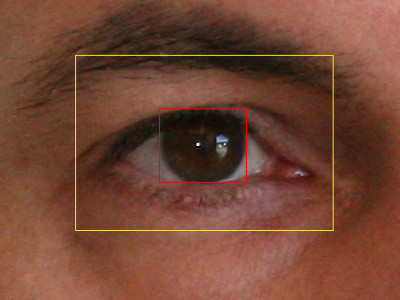}
	}%
	\subfloat[Coarse\label{fig:masks_original:b}]{\includegraphics[width=0.45\columnwidth]{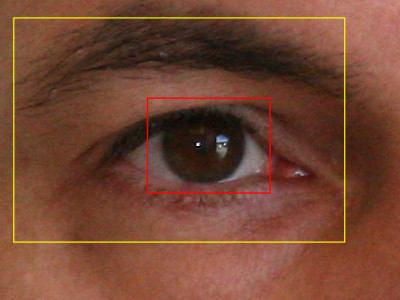}
	}%

	\caption{Examples of fine and coarse annotations of both the iris~(red) and the periocular region~(yellow).}%
	\label{fig:masks_original}%
	\vspace{-1mm}
\end{figure}

In this work, we use coarse annotations both to train and to evaluate our networks.
As can be seen in Fig.~\ref{fig:masks_original}, we define as a coarse annotation the region around the \gls*{roi} so that the edges of the bounding box remain outside the limits of the fine annotations proposed by Severo et al.~\cite{severo2018benchmark}.
More specifically, the delimited region is larger than the one typically used in fine annotations, and the iris is not well-centered.
Also, in some cases, the eyebrows were left out the \gls*{roi}, as the images from some databases used in this work do not contain that~region.

It is worth noting that the coarse annotations were made manually by two volunteers and that no strict rules of how annotations should be made were defined (besides simple instructions and the fact that were coarse and not fine annotations).
Hence, there are random variations (in size, position, aspect ratio, etc.) among annotations of different~images.

We believe coarse annotations can be used in recognition systems based on the iris and/or the periocular region, given the much smaller engineering effort required to manually annotate the training images.
In other words, we conjecture that deep models for person identification may achieve promising results even when these regions are not perfectly~segmented.

\section{Experimental Setup}
\label{sec:experiments}

In this section, we present the databases and also the evaluation protocol used in our experiments. The experiments were carried out on eleven databases, which are described in Section~\ref{sec:datasets}.
Note that we trained/tested the networks on each dataset separately.
All experiments were performed on a computer with an Intel\textsuperscript{\textregistered} Core™ i$7$-$7700$ $4.20$GHz CPU, $16$ GB of RAM and two NVIDIA Titan Xp GPUs.

\subsection{Databases}
\label{sec:datasets}

We employed the following public databases: \interval~\cite{CASIA2010}, \lamp~\cite{CASIA2010}, \thousand~\cite{CASIA2010}, \cross~\cite{Sequeira2016}, \csip~\cite{Santos2015}, \miche~\cite{DeMarsico2015}, \mobbio~\cite{Sequeira2014a}, \niceii~\cite{proenca2010}, \polyu~\cite{Nalla2017}, \ubiris~\cite{proenca2010} and \visob~\cite{Rattani2016}.
An overview of the important features of all databases used in this work can be seen in Table~\ref{tab:overview_datasets}. These databases were chosen because they are widely used in the biometric recognition literature~\cite{zanlorensi2018impact, silva2018multimodal, aginako2016machine,deshpande2014iris}, which we plan to investigate in future works.

\begin{table}[!htb]
	\centering
	\vspace{-1.5mm}
	\caption{Overview of the important features of the databases used in this work. All of these are a subset of the original~database.}
	\label{tab:overview_datasets}
	\vspace{-1.5mm} 
	\resizebox{0.99\columnwidth}{!}{%
	\begin{tabular}{@{}cccccc@{}}
		\toprule
		\textbf{Database} & \textbf{Year} & \textbf{Images} & \textbf{Subjects} & \textbf{Resolution} & \textbf{Wavelength}\\ \midrule
		\interval~\cite{CASIA2010} & $2010$ & $2{,}639$ & $249$ & $320\times280$ & \gls*{nir} \\
		\lamp~\cite{CASIA2010} & $2010$ & $16{,}212$ & $411$ & $640\times480$ & \gls*{nir} \\ 
		\thousand~\cite{CASIA2010} & $2010$ & $20{,}000$ & $1{,}000$ & $640\times480$ & \gls*{nir} \\
		\cross \cite{Sequeira2016} & $2016$ & $1{,}920$ & $120$ & $400\times300$ & \gls*{vis}\\
		\csip* \cite{Santos2015} & $2015$ & $2{,}004$ & $50$ & Various & \gls*{vis} \\
		\miche* \cite{DeMarsico2015} & $2015$ & $3{,}191$ & $92$ & Various & \gls*{vis} \\
		\mobbio \cite{Sequeira2014a} & $2014$ & $1{,}206$ & $105$ & $300\times200$ & \gls*{vis} \\	
		\niceii \cite{proenca2010} & $2010$ & $2{,}000$ & n/a & $400\times300$ & \gls*{vis} \\	
		\polyu \cite{Nalla2017} & $2017$ & $6{,}270$ & $209$ & $640\times480$ & \gls*{vis} \\	
		\ubiris \cite{proenca2010} & $2010$ & $11{,}101$ & $261$ & $400\times300$ & \gls*{vis} \\	
		VISOB* \cite{Rattani2016} & $2016$ & $95{,}046$ & $550$ & Various & \gls*{vis}\\
		\midrule
		\gls*{nir} & & $38{,}851$ & & & \\
		\gls*{vis} & & $122{,}738$ & & & \\
 Total & & $161{,}590$ & & & \\
		\bottomrule \\[-2ex]
		\noindent * Cross-sensor databases
	\end{tabular}
	}
\end{table}

\textbf{\interval}: the iris images of this database were captured with a close-up iris camera developed by the authors themselves.
The database consists of $2{,}639$ images from $249$ subjects and $395$ classes, with a resolution of $320\times280$ pixels, obtained in two~sections.

\textbf{\lamp}: the images were collected using a non-fixed sensor and, thus, the individuals collected the iris image with the sensor in their own hands.
While capturing the images, a lamp was turned on and off in order to produce more intraclass variations due to pupil contraction and expansion, creating a nonlinear deformation.
A total of $16{,}212$ images with a resolution of $640\times480$ pixels from $411$ subjects and $819$ classes were collected in a single~section.

\textbf{\thousand}: this database contains $20{,}000$ iris images from $1000$ subjects with a resolution of $640\times480$ pixels, which were collected in a single section using an IKEMB-$100$ camera.

\textbf{\cross}: this database subset is composed of \gls*{vis} images. Eight images of each eye were captured from $120$ subjects, totaling $1{,}920$ images.
The images have dimensions of $400\times300$ pixels.
All images were obtained at a distance of $1.5$ meters, in an uncontrolled indoor environment, with a wide variation of ethnicity, eye colors, and lighting~conditions.

\textbf{\csip}: this database has images acquired with four different mobile devices: \textit{Sony Ericsson Xperia Arc~S} (rear $3{,}264\times2{,}448$), \textit{iPhone~$4$} (front $640\times480$, rear $2{,}592\times1{,}936$), \textit{THL W$200$} (front $2{,}592\times1{,}936$, rear $3{,}264\times2{,}448$), and \textit{Huawei U$8510$} (front $640\times480$, rear $2{,}048\times1{,}536$).
The database has $2{,}004$ images from $50$ subjects.

\textbf{\miche}: this database contains $3{,}732$ images from $92$ subjects acquired by mobile devices in visible light.
In order to simulate a real application, the iris images were obtained by the users themselves, indoors and outdoors, with and without glasses.
Images of only one eye of each individual were captured.
The mobile devices used and their respective resolutions are the following: \textit{iPhone~$5$} ($1{,}536\times2{,}048$), \textit{Samsung Galaxy S$4$} ($2{,}322\times4{,}128$) and \textit{Samsung Galaxy Tablet II} ($640\times480$).

\textbf{\mobbio}: this database has face, iris, and voice biometric data belonging to $105$ subjects.
The data was acquired with the mobile device \textit{Asus Transformer Pad (TF$300$T)}.
The iris images were obtained in two different lighting conditions, with varying eye orientations and occlusion levels.
For each subject, 16 images (8 of each eye) were captured.

\textbf{\niceii}: this database, a subset of \ubiris, contains $2{,}000$ images with a resolution of $400\times 300$ pixels and was employed in the NICE.II contest. The number of subjects of this set was not directly specified.

\textbf{\polyu}: this database has $6{,}270$ iris images with a resolution of $640\times480$ pixels, with $15$ images of each eye from $209$ subjects obtained in the visible spectrum~\cite{Nalla2017}.

\textbf{\ubiris}: this database contains $11{,}101$ RGB images captured with a Canon EOS $5$D camera and resolution of $400\times 300$ pixels, from $261$ subjects (i.e., $522$ irises)~\cite{proenca2010}.

\textbf{\visob}: front cameras of three mobile devices were used to obtain the images of this database, such as the iPhone~$5$S at $720$p resolution, Samsung Note~$4$ at $1080$p resolution and Oppo~N$1$ also at $1080$p resolution.
The images were captured in $2$ sessions for each of the $2$ visits, which occurred between $2$ and $4$ weeks, totaling $158{,}136$ images from $550$ subjects.

\subsection{Evaluation Protocol}
\label{sub_sec:evaluation_protocol}

The evaluation of an automatic detection approach is performed in a pixel-to-pixel comparison between the ground truth and the predicted bounding boxes.
Therefore, we use the mean $F$-score, \gls*{iou} and \gls*{map} evaluation~metrics.
Following Severo et al.~\cite{severo2018benchmark}, to first compute the precision and recall metrics and then the~$F$-score, we consider as correct the bounding boxes detected with an \gls*{iou} value above~$0.5$ with the ground~truth. 
This bounding box evaluation, defined in the PASCAL VOC Challenge~\cite{everingham2010pascalvoc}, is interesting since it penalizes both over- and under-estimated~objects.

It is worth noting that we use coarse annotations as the ground truth, as the databases do not provide fine annotations of the position of the iris and periocular regions on each image.
In this sense, instead of evaluating the predicted bounding boxes in relation to the exact location of the iris/periocular region, we evaluated how close to the ground truth it is.

In order to perform a fair evaluation and comparison of the proposed approaches, we divided each database into three subsets, being $40\%$ of the images for training, $40\%$ for testing and $20\%$ for validation. 
We adopt this protocol (i.e., with a larger test set) to provide more samples for analysis of statistical significance. 
Also, in the statistical direction, we perform the Wilcoxon signed-rank test~\cite{wilcoxon1970critical} to verify if there is a statistical difference between the detection~approaches.
\begin{table*}[!htb]
    \vspace{-2mm}
	\centering	
    \caption{Detection results. The Single and Multi columns present the results obtained when detecting the iris and  periocular regions separately and simultaneously, respectively. The values in bold represent the highest \gls*{iou} values obtained, while the highlighted results indicate the cases in which there is no statistical difference according to the Wilcoxon statistical~tests.}
	\label{tab:results}
	\vspace{-1.5mm}
	\resizebox{0.975\linewidth}{!}{
		\begin{tabular}{c||cc|cc||cc|cc||cc|cc}
			\toprule
			
			\multicolumn{1}{c}{\multirow{3}{*}{\textbf{Database}}}       & \multicolumn{4}{c}{\textbf{F-score}} & \multicolumn{4}{c}{\textbf{\gls*{iou} (\%)}} &  \multicolumn{4}{c}{\textbf{\gls*{map} (\%)}} \\
			\multicolumn{1}{c}{}                & \multicolumn{2}{c|}{\textbf{\yolo}}    & \multicolumn{2}{c||}{\textbf{\faster}} & \multicolumn{2}{c|}{\textbf{\yolo}} & \multicolumn{2}{c|||}{\textbf{\faster}} & \multicolumn{2}{c|}{\textbf{\yolo}} & \multicolumn{2}{c}{\textbf{\faster}}\\
			\multicolumn{1}{c}{} & \textbf{Multi}   & \textbf{Single} & \textbf{Multi}  & \textbf{Single}  & \textbf{Multi}  & \textbf{Single}  & \textbf{Multi}  & \textbf{Single}  & \textbf{Multi}   & \textbf{Single} & \textbf{Multi}  & \textbf{Single}\\  \midrule
            \multicolumn{13}{c}{\textbf{Iris}} \\
            \midrule
			\textbf{\interval} & $0.90$   & $0.92$ & $0.97$ & $0.96$ & $82.81$          & $86.20$ & $\textbf{94.77}                           $&  $ 93.98$                                  & 100.00   & 100.00   & 100.00   & 93.98  \\
			\textbf{\lamp}     & $0.96$   & $0.96$ & $0.98$ & $0.95$ & $92.38$          & $93.06$ & \cellcolor[HTML]{C0C0C0}$96.08$  & \cellcolor[HTML]{C0C0C0}$\textbf{97.31}$           & 99.98 & 99.98 & 99.73 & 97.31  \\
			\textbf{\thousand} & $0.97$   & $0.98$ & $0.98$ & $0.98$ & $95.71$          & $94.39$ & $\textbf{97.72}$                           &  $97.58$                                   & 99.96 & 99.97 & 99.65 & 97.58  \\
			\textbf{\cross}    & $0.92$   & $0.92$ & $0.94$ & $0.94$ & $85.79$          & $86.45$ &\cellcolor[HTML]{C0C0C0}$90.39$            & \cellcolor[HTML]{C0C0C0}$\textbf{90.44}$  & 100.00   & 100.00   & 100.00   & 90.44  \\
			\textbf{\csip}     & $0.92$   & $0.73$ & $0.95$ & $0.95$ & $87.97$          & $58.12$ & \cellcolor[HTML]{C0C0C0}$\textbf{91.61}$  & \cellcolor[HTML]{C0C0C0}$91.55$           & 98.68 & 98.69 & 100.00   & 91.55  \\
			\textbf{\miche}    & $0.88$   & $0.83$ & $0.92$ & $0.92$ & $80.32$          & $72.07$ & $86.27$                                    & $\textbf{86.48}$                           & 97.39 & 94.32 & 100.00   & 92.48  \\
			\textbf{\mobbio}   & $0.95$   & $0.95$ & $0.96$ & $0.96$ & $91.52$          & $91.40$ & \cellcolor[HTML]{C0C0C0}$\textbf{94.14}$ & \cellcolor[HTML]{C0C0C0}$93.79$           & 100.00   & 100.00   & 100.00   & 93.79  \\
			\textbf{\niceii}   & $0.90$   & $0.91$ & $0.93$ & $0.82$ & $83.39$          & $84.83$ & $\textbf{88.41}$                           & $78.20$                                    & 98.92 & 99.32 & 99.32 & 78.20  \\
			\textbf{\polyu}    & $0.91$   & $0.86$ & $0.94$ & $0.94$ & $\textbf{93.81}$ & $76.32$ & $89.12$                                    & $89.31$                                    & 99.74 & 93.79 & 100.00   & 89.31  \\
			\textbf{\ubiris}   & $0.89$   & $0.89$ & $0.91$ & $0.91$ & $81.16$          & $81.75$ & $85.16$                                    & $\textbf{85.26}$                           & 99.35 & 99.00 & 100.00   & 85.29  \\
			\textbf{\visob}    & $0.91$   & $0.89$ & $0.96$ & $0.96$ & $85.04$          & $81.32$ & $\textbf{93.09}$                           & $92.80$                                    & 99.53 & 99.34 & 99.90 & 92.80  \\
			\midrule
			\multicolumn{13}{c}{\textbf{Periocular Region}}\\
			\midrule
			\textbf{\interval} & $0.96$   & $0.98$ & $0.98$ & $0.98$ & $92.65$          & $96.19$ & $\textbf{97.80}$                                   & $96.79$                            & 98.62 & 100.00   & 100.00   &  97.80 \\
			\textbf{\lamp}     & $0.98$   & $0.97$ & $0.99$ & $0.98$ & $97.15$          & $96.02$ & $\textbf{98.08}$                          & $97.71$                                     & 99.95 & 99.95 & 99.97 &  97.70 \\
			\textbf{\thousand} & $0.97$   & $0.98$ & $0.99$ & $0.99$ & $95.92$          & $96.44$ & \cellcolor[HTML]{C0C0C0}$\textbf{98.19}$ & \cellcolor[HTML]{C0C0C0}$98.19$           & 99.89 & 99.94 & 99.97 &  98.18 \\
			\textbf{\cross}    & $0.92$   & $0.92$ & $0.96$ & $0.96$ & $86.86$          & $86.89$ & \cellcolor[HTML]{C0C0C0}$\textbf{92.74}$ & \cellcolor[HTML]{C0C0C0}$92.56$           & 97.84 & 99.66 & 100.00   &  92.56 \\
			\textbf{\csip}     & $0.95$   & $0.95$ & $0.87$ & $0.96$ & $91.61$          & $91.76$ & $84.97$                                   &  $\textbf{92.96}$                           & 99.83 & 100.00   & 83.61 &  92.96 \\
			\textbf{\miche}    & $0.85$   & $0.85$ & $0.90$ & $0.90$ & $75.88$          & $74.97$ & $\textbf{83.66}$                          & $83.51$                                     & 93.82 & 96.33 & 98.77 &  93.51 \\
			\textbf{\mobbio}   & $0.96$   & $0.96$ & $0.97$ & $0.97$ & $94.21$          & $94.09$ & $\textbf{95.50}$                          & $94.83$                                     & 100.00   & 100.00   & 100.00   &  94.83 \\
			\textbf{\niceii}   & $0.88$   & $0.90$ & $0.92$ & $0.92$ & $80.52$          & $82.44$ & \cellcolor[HTML]{C0C0C0}$\textbf{86.91}$ & \cellcolor[HTML]{C0C0C0}$86.66$            & 97.23 & 99.55 & 99.76 &  86.66 \\ 
			\textbf{\polyu}    & $0.96$   & $0.87$ & $0.98$ & $0.98$ & $93.57$          & $77.95$ & $\textbf{96.74}$                          & $96.41$                                     & 99.48 & 99.56 & 100.00   &  96.41 \\
			\textbf{\ubiris}   & $0.87$   & $0.88$ & $0.91$ & $0.91$ & $78.98$          & $80.03$ & $85.19$                                   & $\textbf{85.44}$                            & 83.12 & 98.35 & 99.64 &  85.44 \\
			\textbf{\visob}    & $0.93$   & $0.94$ & $0.97$ & $0.98$ & $87.17$          & $89.11$ & $96.08$                                   & $\textbf{96.35}$                            & 95.64 & 99.98 & 99.83 &  96.35 \\

			\bottomrule
		\end{tabular}
	}		
	\vspace{-1mm}
\end{table*}

\section{Results}
\label{sec:results}

The experiments were carried out using the protocol presented in Section~\ref{sub_sec:evaluation_protocol}. To compare the proposed approaches, we report the $F$-score values in order to analyze the trade-off between precision and recall measures, however, we focus on the \gls*{iou}~metric since we want to assess how close are the predicted bounding boxes compared to the ground truth. 

When analyzing the results regarding \textbf{\textit{iris}} detection (see top of Table~\ref{tab:results}), in $10$ of $11$ experiments the highest mean \gls*{iou} value was achieved using \faster.
In general, the best results were obtained when simultaneously detecting the iris and the periocular region.
The exceptions are in the \lamp, \cross, \miche, and \ubiris databases, where detecting both regions separately performed better, probably due to the fact that there are not many variations in iris and periocular region arrangement in the images of these databases.
However, as the difference in the results obtained with both approaches is very small, we applied the Wilcoxon signed-rank test and observed that there is no statistical difference between detecting the iris and the periocular region simultaneously or separately in the \lamp, \cross, \csip and \mobbio databases. 
In this way, in Table~\ref{tab:results}, we highlighted (light gray) the results obtained in these~databases.

Similar behavior occurred in the detection of the \textbf{\textit{periocular}} region, however, in this case, all the best results were attained employing the \faster model. 
In this scenario, the detection results using the single-class detection approach \csip, \ubiris and \visob databases presented the best values.
Similar to the results on iris detection, the difference between the \gls*{iou} values attained between the approaches is close and there is no statistical difference in the \thousand, \cross and \niceii databases and that result was also highlighted in Table~\ref{tab:results}.

We emphasize that most of the best results were obtained using the \faster + \gls*{fpn} approach, which we believe to be justified by the fact that \glspl*{fpn} perform a better feature map extraction compared to other approaches~\cite{FPN}. 

\begin{figure}[!htb]
    \vspace{-3.5mm}
	\centering
	\subfloat[Periocular Region \label{fig:a}]{\includegraphics[width=0.43\columnwidth]{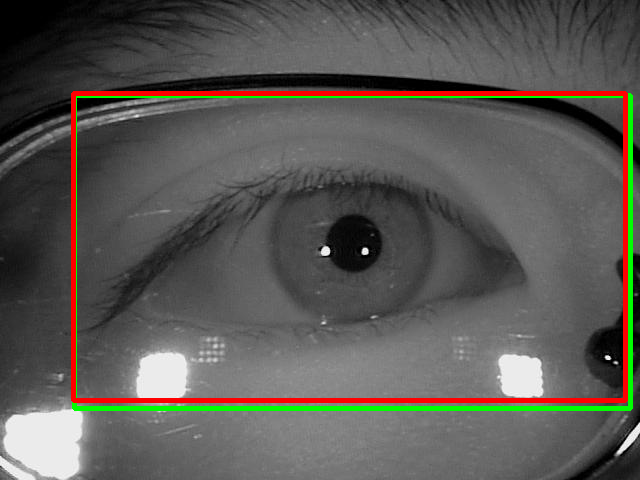}
	} \,
	\subfloat[Iris Detection \label{fig:b}]{\includegraphics[width=0.43\columnwidth]{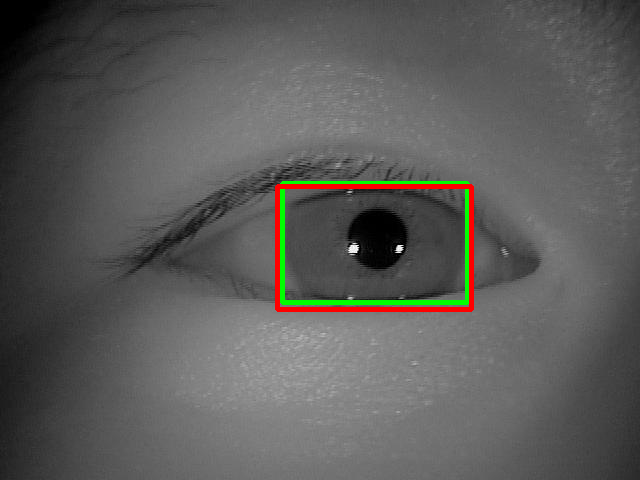}
	}%
	
	\caption{Best iris and periocular region detection performed by the \faster simultaneous detection approach. The green bounding boxes represent the coarse annotations, while the red ones represent the detected~regions.}
	\label{fig:best}
\end{figure}

It should be noted that the \gls*{iou} values obtained were higher than $95$\% for both iris and periocular region detection in the databases where the images were captured using a \gls*{nir} sensor.
These results were achieved by using the \faster simultaneous detection approach, and the better detected iris and periocular region can be seen in Figure~\ref{fig:best}.

Despite the good results it is necessary observe that in the databases in which the images were captured using more than one sensor, that there was no preprocessing of the image (i.e., \miche and \csip) or that composed with lower quality images (i.e., \ubiris and \niceii) we obtained results with \gls*{iou} values lower than $90$\% when detecting both the iris and periocular regions simultaneously.
By analyzing these images, we can understand what made the results obtained by the approaches on these databases below than $90$\% of \gls*{iou}:
i)~the use of eyeglasses;
ii)~the presence of more than one eye;

\section{Conclusions}
\label{sec:conclusions}

In this work, we compared the detection of the iris and the periocular region being performed separately or simultaneously using two well-known object detectors, observing a better performance of the \faster + \gls*{fpn}~approach.

The detection of both regions being performed simultaneously produced better results in most databases, for both the iris and the periocular region. This leads us to believe that using this approach gives the neural network a certain understanding of the context present in the image.

We also coarsely labeled \textbf{$161{,}590$} images for iris and periocular region detection. These annotations are publicly available to the research community, assisting the development and evaluation of new detection approaches as well as the fair comparison among published~works.

There is still room for improvements in the simultaneous detection of iris and periocular region. As future work, we intend to (i)~design new and better network architectures; (ii)~design a general and independent sensor approach, where the image sensor is first classified and then the iris and the periocular region are simultaneously detected with a specific approach; (iii)~compare the proposed approach with methods applied in other domains; (iv)~create a context-aware object-detection~architecture; and (v)~design a cascade detection approach for iris and periocular region detection.
\balance
\section*{Acknowledgements}
This work was supported by the National Council for Scientific and Technological Development~(CNPq) (grant numbers~428333/2016-8 and 313423/2017-2) and the Coordination for the Improvement of Higher Education Personnel~(CAPES) (Social Demand Program).
The Titan~Xp GPUs used for this research were donated by the NVIDIA~Corporation.


\bibliographystyle{IEEEtran}
\bibliography{bibtex}

\end{document}